\documentclass{article}

\usepackage[sc]{mathpazo} 
\usepackage[T1]{fontenc} 
\usepackage{microtype} 

\usepackage[hmarginratio=1:1,top=32mm,columnsep=20pt]{geometry} 
\usepackage{multicol} 
\usepackage[hang, small,labelfont=bf,up,textfont=it,up]{caption} 
\usepackage{booktabs} 
\usepackage{float} 
\usepackage{hyperref} 

\usepackage{lettrine} 
\usepackage{paralist} 

\usepackage{abstract} 

\usepackage{titlesec} 
\renewcommand\thesection{\Roman{section}} 
\renewcommand\thesubsection{\Roman{subsection}} 
\titleformat{\section}[block]{\large\scshape\centering}{\thesection.}{1em}{} 
\titleformat{\subsection}[block]{\large}{\thesubsection.}{1em}{} 

\usepackage{graphicx}
\usepackage{framed,multirow}
\usepackage{amssymb}
\usepackage{latexsym}
\usepackage{url}
\usepackage{xcolor}
\definecolor{newcolor}{rgb}{.8,.349,.1}
\usepackage{amsmath}
\usepackage{float}
\usepackage{booktabs}
\usepackage{siunitx}
\usepackage[caption = false]{subfig}
\usepackage{epstopdf}
\usepackage{enumitem}
\usepackage{textcomp}
\usepackage{natbib}

\title{\vspace{-15mm}\fontsize{24pt}{10pt}\selectfont\textbf{Feature Selection via Binary Simultaneous Perturbation Stochastic Approximation}} 

\author{
\large
\textsc{Vural Aksakalli}\thanks{Corresponding Author: \href{mailto:aksakalli@sehir.edu.tr}{aksakalli@sehir.edu.tr}}\\[2mm]  \textsc{Milad Malekipirbazari}\\[2mm] 
\normalsize Department of Industrial Engineering, Istanbul Sehir University, 34662, Istanbul, Turkey \\ 
\vspace{-5mm}
}
\date{}

\begin{document}

\maketitle 


\begin{abstract}

\noindent 
Feature selection (FS) has become an indispensable task in dealing with today's highly complex pattern recognition problems with massive number of features. In this study, we propose a new wrapper approach for FS based on binary simultaneous perturbation stochastic approximation (BSPSA). This pseudo-gradient descent stochastic algorithm starts with an initial feature vector and moves toward the optimal feature vector via successive iterations. In each iteration, the current feature vector's individual components are perturbed simultaneously by random offsets from a qualified probability distribution. We present computational experiments on datasets with numbers of features ranging from a few dozens to thousands using three widely-used classifiers as wrappers: nearest neighbor, decision tree, and linear support vector machine. We compare our methodology against the full set of features as well as a binary genetic algorithm and sequential FS methods using cross-validated classification error rate and AUC as the performance criteria. Our results indicate that features selected by BSPSA compare favorably to alternative methods in general and BSPSA can yield superior feature sets for datasets with tens of thousands of features by examining an extremely small fraction of the solution space. We are not aware of any other wrapper FS methods that are computationally feasible with good convergence properties for such large datasets.
\end{abstract}


\section{Introduction}
\label{sec:intro}

\lettrine[nindent=0em,lines=3]{R}ecent emergence of datasets with massive numbers of features has made pattern recognition an ever-challenging task. In particular, such high numbers of features give rise to various issues such as (1) overfitting, poor generalization, and inferior prediction performance, (2) slow and computationally expensive predictors, and (3)  difficulty in comprehending the underlying process. Feature selection (FS) can be defined as selecting a subset of available features in a dataset that are associated with the response variable by excluding irrelevant and redundant features. An effective feature selection process mitigates the problems associated with large datasets in the sense that it results in (1) better classification performance, (2) reduced storage and computational cost, and (3) generalized and more interpretable models. An alternative to FS for dimensionality reduction is feature extraction (FE) wherein original features are first combined and then projected into a new feature space with lower dimensionality. A major downside of FE is that the transformed features lose their physical meaning, which complicates further analysis of the model and makes it difficult to interpret. Thus, FS is superior to FE in terms of readability and interpretability~\citep{TangAL14}.

In the FS problem, the goal is to identify the optimal subset of the available set of features with respect to a particular classification performance criterion. For a dataset with $p$ features, the size of the solution space is $2^p-1$, which makes the FS problem computationally intractable. FS methods currently available in the literature fall into three broad categories: filters, wrappers, and embedded methods. Filter methods utilize statistical characteristics of data in order to remove poorly associated features. Filter methods ignore the effects of the chosen feature set on the performance of the intended classifier. Wrapper methods alleviate this issue by exploring the space of feature subsets for the set that optimizes a desired performance criterion for a given classifier. Embedded methods' performance criterion is the same as that of wrappers, yet they incorporate the FS process as part of the training process~\citep{TangAL14}.    

Wrapper FS algorithms can be divided into four categories based on their search strategies: complete, heuristic, meta-heuristic, and search with artificial neural networks. Complete search methods are clearly infeasible for a large number of available features. Heuristic wrapper methods include greedy hill climbing, branch and bound techniques, beam search, and best first algorithms. Two popular greedy hill climbing algorithms are Sequential Forward Selection (SFS) and Sequential Backward Selection (SBS). The former starts with an empty set and adds informative features to the feature set one by one, while the latter starts with the full set and removes one irrelevant or redundant feature at each step. These two methods do not consider re-evaluation of the omitted features, which causes a nesting effect. In order to overcome this issue, Sequential Forward Floating Selection (SFFS) and Sequential Backward Floating Selection (SBFS) were proposed~\citep{PudilNK94}. In SFFS, for instance, whenever a feature is added to the set, the current feature set is scanned and features whose removal would improve the performance criterion are removed from the set. This process is continued until no new features can be added that improve the performance criterion. 

Various meta-heuristic approaches have been proposed for feature selection, including genetic algorithms (GA)~\citep{raymer2000dimensionality, tsai2013genetic, OluleyeALD14}, simulated annealing~\citep{DebuseR97}, ant colony optimization~\citep{Al05}, particle swarm optimization~\citep{WangYTXJ07}, and tabu search~\citep{TahirBK07}. Such algorithms are random in nature and they can outperform not only complete search but also heuristic sequential FS methods. Nonetheless, these algorithms typically have high computational requirements and, for the most part, there are no optimality guarantees for the feature subset they find. Artificial Neural Networks (ANN) have also been considered for feature selection due to their ability to discover hidden patterns under noisy conditions. The drawback of this method is the time consuming training process, which necessitates incorporation of other techniques to decrease the total execution time~\citep{LedesmaC08}. 

The purpose of this study is to introduce a new wrapper approach for FS  based on a pseudo-gradient descent stochastic optimization algorithm called Binary Simultaneous Perturbation Stochastic Approximation (BSPSA). This algorithm starts with a random solution vector and moves toward the optimal solution vector via successive iterations in which the current solution vector's individual components are perturbed simultaneously by random offsets from a qualified probability distribution. Regarding gradient descent based optimization approaches for FS, \citet{gadat2007stochastic} discuss a stochastic gradient descent algorithm where a probability distribution is first estimated on the full set of features whose mass is then distributed over the more informative features. The closest work to ours is that of~\citet{JohannsenWSP04} wherein the authors use conventional (continuous-space) SPSA for feature selection for a nearest neighbor classifier with the Minkowski distance metric for two particular datasets. On the other hand, binary SPSA takes advantage of the inherent binary nature of FS and eliminates the need for defining and fine-tuning additional algorithm parameters as in continuous SPSA.
 
We present computational experiments on datasets with numbers of features ranging from a few dozens to thousands using three widely-used classifiers as wrappers: nearest neighbor, decision tree (C4.5), and linear support vector machine (SVM). We compare our methodology against the full set of features as well as other popular FS methods including binary GA, SFS, SBS, and SFFS. In our computational experiments, we benchmark BSPSA against the continuous SPSA implementation of~\citet{JohannsenWSP04} as well as the binary GA implementation of~\citet{OluleyeALD14}. We use average cross-validation error as well as AUC as the performance criteria. To our knowledge, the wrapper FS problem is the first application of BSPSA in the literature. In addition, we are not aware of any other wrapper FS methods that are computationally feasible for datasets with thousands of features with good convergence properties.

\section{SPSA and Binary SPSA}
\label{sec:spsa}

Optimization has played a central role in pattern recognition due to our never-ending quest for highest possible classifier performance. A popular deterministic method for finding a local minimum of a real-valued objective function $L: \mathbb{R}^p \to \mathbb{R}$ is the gradient descent approach. This method makes use of the simple fact that steepest descent in the objective function value at a given point occurs in the negative direction of the function's gradient at that point, i.e., the vector of its first partial derivatives. The objective function is sometimes referred to as ``loss'' function in the case of minimization, which is the convention in this work. The gradient descent algorithm starts with an initial guess for the solution, evaluates the function's gradient, and moves in the direction of negative gradient in an amount specified by a step size function. The algorithm continues in this fashion in an attempt to converge to a locally optimal point where the gradient is zero. 

A fundamental assumption in conventional gradient descent algorithms is that the loss function (and its derivatives') information is explicitly available. Hence, these algorithms are not directly applicable in situations where the loss function is not known explicitly and/or it can be observed only via noisy measurements. Such cases are especially relevant within the context of pattern recognition due to the fact that common classification performance criteria such as cross-validation error rates are inherently noisy and there are no explicit functional forms for them that can be used for proper gradient evaluation.

Situations such as above give rise to stochastic pseudo-gradient descent algorithms that approximate the gradient from noisy loss function measurements. In particular, these algorithms do not require detailed modeling information between the loss function and the variables to be optimized. In addition, such algorithms formally account for the noise in function measurements. 

The classical stochastic optimization approach in the absence of the explicit loss function information is the Kiefer-Wolfowitz finite-difference stochastic approximation (FDSA)~\citep{KieferW52}. The gradient approximation employed in FDSA is the finite difference where the number of loss function measurements required at each iteration for gradient estimation is $2p$ where $p$ is the size of the solution vector. Introduced by~\citet{Spall92}, SPSA makes a major improvement to FDSA by providing the same level of statistical accuracy with only two measurements to form one gradient approximation, resulting in a dramatic $p-$fold decrease in execution time. This gradient approximation is obtained by simultaneous perturbation to the solution vector's individual components at each iteration as described below. 

\subsection{SPSA}

Let $L(w): \mathbb{R}^p \to \mathbb{R}$ denote the loss function to be optimized where an explicit functional form for $L$ is not available, yet one can make noisy measurements $y(w):= L(w) + \epsilon(w)$ where $\epsilon$ denotes noise. The gradient of $L$ is defined as
\begin{equation}
g(w) := \frac{\partial L}{\partial w}.
\end{equation}

As in any typical gradient descent based optimization algorithm, SPSA starts with an initial estimate $\hat{w}_0$ and iterates in accordance with the recursion below to find a local minimizer $w^*$:
\begin{equation}\label{eq:scheme}
\hat{w}_{k+1} := \hat{w}_{k} - a_k \hat{g}_k(\hat{w}_k).
\end{equation}
\noindent

Here, $a_k$ is a nonnegative iteration gain sequence and $\hat{g}_k(\hat{w}_k)$ denotes the approximate gradient at $\hat{w}_k$. Since it is assumed that $L$ is not known explicitly, the gradient $g(w)$ is not readily available and thus it must be approximated. For simplicity, suppose $w$ is scalar. For this approximation, SPSA goes back to the basics and makes use of the definition of the gradient of $L$ at $w$, which is the slope of the tangent line to  $L$ at $w$. SPSA ``perturbs" the current iterate $w$ by a small amount in each direction as $w + \delta$ and $w - \delta$ where $\delta > 0$. SPSA then approximates $g(w)$ as the slope of the secant line whose end points are $w + \delta$ and $w - \delta$ respectively. Since $L$ can only be observed via noisy measurements, $L(w + \delta)$ and $L(w - \delta)$ are also approximated as $y(w + \delta)$ and $y(w - \delta)$ respectively. 

In SPSA, the perturbation amount $\delta$ is taken as $c_k \Delta_k$ where $c_k$ is a nonnegative gradient gain sequence and $\Delta_k$ is the $p-$dimensional simultaneous perturbation vector. SPSA imposes certain regularity conditions on $\Delta_k$~\citep{Spall92}. In particular, each component of $\Delta_k$ needs to be generated independently from a symmetric zero mean probability distribution with a finite inverse, such as the symmetric Bernoulli distribution (e.g., $+1$ or $-1$ with 0.5 probability). Due to the finite inverse requirement, for instance, the uniform and normal distributions are not allowed. Simultaneous perturbations around the current iterate $\hat{w}_{k}$ are then defined as 
\begin{equation}
\hat{w}_k^{\pm} := \hat{w}_k \pm c_k \Delta_k.
\label{eq:w_pm}
\end{equation}

Once $y(\hat{w}_{k}^{+})$ and $y(\hat{w}_{k}^{-})$ are measured, the estimate of gradient $\hat{g}_k$ is computed as:
\begin{equation}
\hat{g}_k(\hat{w}_k) := \frac{y(\hat{w}_{k}^{+})-y(\hat{w}_{k}^{-})}{2c_k} \begin{bmatrix}
\Delta_{k1}^{-1}\\
\Delta_{k2}^{-1}\\
\vdots\\
\Delta_{kp}^{-1}\\
\end{bmatrix}.
\label{eq:est_grad}
\end{equation}
\noindent

Observe that SPSA requires three loss function evaluations in each iteration: $y(\hat{w}_k^{+})$,  $y(\hat{w}_k^{-})$, and $y(\hat{w}_{k+1})$. The first two evaluations are used to approximate the gradient and the third one is used to measure the performance of the subsequent iterate, i.e., $\hat{w}_{k+1}$.

The iteration gain sequence is specified as $a_k := a/(A+k)^\alpha$ where $A$ is the stability constant and the gradient gain sequence is taken as $c_k := c/k^\gamma$. In SPSA, $a,c,A,\alpha,\gamma$ are pre-defined parameters whose proper fine-tuning is critical for satisfactory algorithm performance.

Automatic stopping rules are not available for SPSA-like stochastic approximation algorithms in general. Thus, a typical stopping criterion for SPSA is a pre-specified maximum number of iterations, perhaps in conjunction with a stall limit. Under some mild conditions, SPSA has been shown to converge to a local minimizer almost surely~\citep{Spall92}.

\subsection{Binary SPSA}

\cite{WangS11} discusses a discrete version of SPSA where $w \in \mathbb{Z}^p$ as well as a binary version of SPSA (BSPSA)  as a special case of discrete SPSA. Specifically, suppose $L: \{0,1\}^p \to \mathbb{R}$ and $y$ is a noisy measurement of $L$. In this particular case, estimate of the gradient, $\hat{g}_k$, is computed as:
\begin{equation}
\hat{g}_k(\hat{w}_k) = \frac{y(R(B(\hat{w}_k^{+})))-y(R(B(\hat{w}_k^{-})))}{2c} \begin{bmatrix}
\Delta_{k1}^{-1}\\
\Delta_{k2}^{-1}\\
\vdots\\
\Delta_{kp}^{-1}\\
\end{bmatrix}.
\label{eq:est_grad_bspsa}
\end{equation}

\noindent
where $R$ is the component-wise rounding operator, $B$ is the component-wise (0,1) bounding operator, $c$ is a positive constant that adjusts the magnitude of the perturbation, and $\hat{w}^{\pm}$ are the random perturbations defined as in Equation~\ref{eq:w_pm}. Observe that one difference of BSPSA from conventional (continuous) SPSA is that the gain sequence $c_k$ is constant, i.e., $c_k=c$. A generic implementation of BSPSA is given below.

\begin{enumerate}[label=\itshape Step \arabic*:]
\setcounter{enumi}{-1}
\setlength{\itemindent}{15pt}
\item Choose an estimate for the initial solution vector $\hat{w}_0$.  
\item Generate $\Delta_{k}$, e.g., from the symmetric Bernoulli distribution with values $\pm 1$. 
\item Compute $\hat{w}_k^{+}$ and $\hat{w}_k^{-}$.
\item Compute $B\big(\hat{w}_{k}^{\pm}\big)$ and then $R\big(\hat{w}_{k}^{\pm}\big)$.
\item After bounding and rounding $\hat{w}_{k}^{\pm}$, evaluate $y(\hat{w}_{k}^{+})$ and $y(\hat{w}_{k}^{-})$.
\item Compute the gradient estimate $\hat{g}_k(\hat{w}_k)$ using Equation~\ref{eq:est_grad_bspsa}.
\item Update the estimate according to the recursion in Equation~\ref{eq:scheme}.
\item Terminate when the maximum number of iterations or the stall limit is reached and report the best solution vector found thus far as the algorithm's output.
\end{enumerate}

\subsection{The FS Problem and BSPSA}

We now give a formal definition of the wrapper feature selection (FS) problem before illustrating how BSPSA can be used for its approximate solution. Let $D=(\textbf{X},Y)$ be a dataset with $p$ features and $n$ observations with $n \times p$ feature data matrix $\textbf{X}$ and $n \times 1$ response vector $Y$. Let the set $X=\{X_1, \ldots, X_p\}$ denote the feature set where $X_i$ denotes the $i-$th feature in $\textbf{X}$. For a classification task at hand, using all the available features might result in overfitting and therefore it is often desirable to identify the subset of features that optimizes a certain classification performance criterion. For a nonempty feature subset $X' \subseteq X$, let $L_C(X',Y)$ denote the true value of the performance criterion associated with wrapper classifier $C$ on $D$, such as the $k-$fold cross-validated classification error rate. Note that $L$ is defined over the space of all possible $k-$fold cross-validations on $D$ (here, $k$ is usually taken as 5 or 10). It is computationally infeasible to compute $L$ for most realistic datasets due to the extremely large number of all possible ways of $k-$fold partitioning of the original dataset. However, we can perform one random cross-validation on $D$ and compute the error rate, which we may denote by $y_C(X',Y)$ as it is essentially a noisy measurement of the classifier $C$'s true error rate $L_C(X',Y)$. That is, $y_C = L_C + \epsilon$. For robustness, one might as well measure the $k-$fold cross-validated classification error rate randomly several times and take the average. The wrapper FS problem is defined as finding the nonempty feature subset $X^*$ such that 

\begin{equation}
X^* := \arg \min_{\substack{ X' \subseteq X \\ X' \neq \emptyset}} y_C(X',Y).
\label{eq:fs_defn}
\end{equation}

\begin{figure}
\centering
\includegraphics[width=1.1\textwidth]{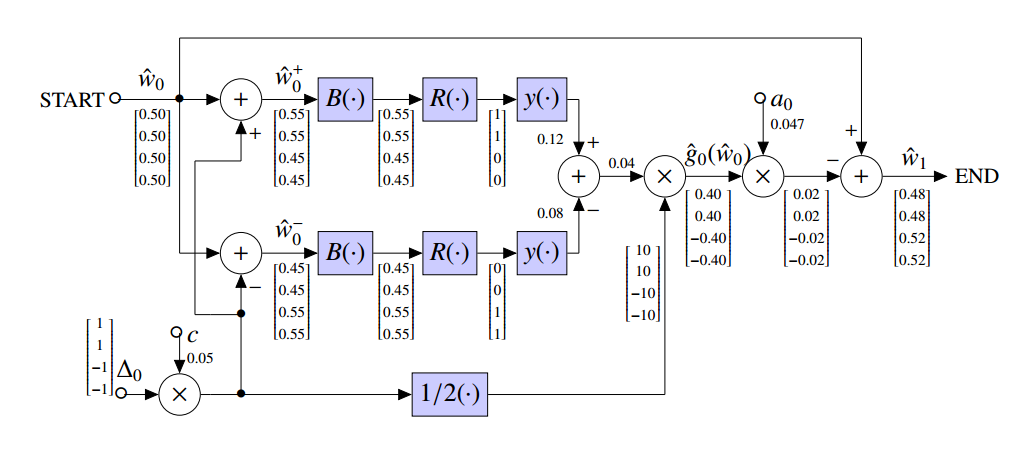}
\caption{Block diagram of the first iteration of BSPSA for a hypothetical dataset with four features}
\label{fig:block}
\end{figure}

In this study, we empirically show that the above optimization problem, which entails evaluation of the true loss function $L$ via the noisy measurements of $y$, can be efficiently and effectively solved via BSPSA. We now present a simple illustration of BSPSA as a wrapper FS method on a hypothetical dataset with four features. We note that this particular example coincides with our specific BSPSA implementation. Figure~\ref{fig:block} shows a block diagram of the algorithm's first iteration. Here, $y$ is assumed to be a cross-validated error rate with respect to a particular classifier. Suppose $c=0.05$, $a=0.75$, $A=100$, and $\alpha=0.6$. Details of this iteration are given below.

\begin{enumerate}[label=\itshape Step \arabic*:]
\setcounter{enumi}{-1}
\setlength{\itemindent}{15pt}
\item Choose the initial solution vector $\hat{w}_0$ as $[0.5,0.5,0.5,0.5]$.  
\item Sample $\Delta_{0}$ from the symmetric Bernoulli distribution as $[1,1,-1,-1]$.
\item Set $\hat{w}_0^{+} = \hat{w}_0 + (0.05)\Delta_{0} = [0.55, 0.55, 0.45, 0.45]$ and $\hat{w}_0^{-} = \hat{w}_0 - (0.05)\Delta_{0} = [0.45, 0.45, 0.55, 0.55]$.
\item Observe that $B(\hat{w}_k^{\pm}) = \hat{w}_k^{\pm}$, and $R(\hat{w}_k^{+}) = [1,1,0,0]$ and $R(\hat{w}_k^{-}) = [0,0,1,1]$. 
\item Evaluate $y([1,1,0,0])$ and $y([0,0,1,1])$, which are assumed to be 0.12 and 0.08 respectively. 

\item Note that $\xi: = 2 (0.05) \ \Delta_{0} = [0.1,0.1,-0.1,-0.1]$. Compute the gradient estimate $\hat{g}_0(\hat{w}_0) = (0.12-0.08)/ \xi = [0.4, 0.4, -0.4, -0.4]$.
\item Observe that $a_0 = 0.75/101^{0.6} = 0.047$. Compute $\hat{w}_{1} = [0.5,0.5,0.5,0.5] - (0.047) [0.4, 0.4, -0.4, -0.4] = [0.48, 0.48, 0.52, 0.52]$. 
\end{enumerate}

Remark that, after rounding, the solution vector $\hat{w}_1$ corresponds to the feature vector $[0,0,1,1]$. Nonetheless, with a slight abuse of terminology, the terms ``solution vector'' and ``feature vector'' shall be used interchangeably in this work whenever no particular distinction needs to be made between the two.

\section{Computational Experiments}
\label{sec:exprmnt}

This section presents computational experiments in order to empirically assess relative performance of BSPSA as a feature selection method using three popular classifiers as wrappers. We experiment with two different sets of datasets: 5 ``small'' datasets with less than 100 features and 5 ``large'' datasets with more than 100 features. We compare BSPSA against the following:

\begin{itemize}
\item The full set of features.
\item The (continuous) SPSA implementation of~\citet{JohannsenWSP04}, which shall be referred to as CSPSA in the rest of this manuscript.
\item Binary Genetic Algorithm (BGA). The particular implementation we use is that of~\citet{OluleyeALD14}.
\item Conventional wrapper methods SFS, SBS, and SFFS. These methods are included only in the case of small datasets due to their extremely high computational requirements for the large ones.
\end{itemize}

Each FS method was executed only once for each dataset. In what follows, we describe the datasets and classifiers used, explain the objective function we consider, and discuss the parameter fine-tuning process for BSPSA.

\subsection{Datasets}

The 10 benchmark datasets we use are from the UCI machine learning~\citep{Lichman13} and ASU feature selection repositories~\citep{ASU}. They are from several different application domains including life, physical, imaging, and microarray fields. These datasets range from 18 to more than 22,000 features and they are described in Table~\ref{tab:datasets}.

\begin{table}[t]
\fontsize{8}{11pt}\selectfont
\begin{center}
\caption{Description of benchmark datasets}\label{tab:datasets}
\begin{tabular}{l l l l l l}
\hline
Name                & Instances     & Features	& Classes   &   Domain		& Source		\\
\hline
Small Datasets		&				&			&			&				&				\\
Ionosphere          & 351       	& 34        & 2         &	Physical	& UCI			\\
Segmentation		& 2310			& 18		& 7			&	Image		& UCI			\\
Sonar		        & 208       	& 60        & 2         &	Image		& UCI			\\
Vehicle			    & 846        	& 18        & 4         &	Image		& UCI			\\
Waveform	        & 5000      	& 21        & 3         &	Physical	& UCI			\\
\hline
Large Datasets		&				&			&			&				&				\\
Arrhythmia          & 452			& 279       & 13        &	Life		& UCI			\\
AR10P		        & 130       	& 2400      & 10        &	Image		& ASU			\\
CLL-SUB-111         & 111       	& 11340     & 3         &	Microarray	& ASU			\\
GLI-85		        & 85        	& 22283     & 2         &	Microarray	& ASU			\\
TOX-171		        & 171       	& 5748      & 4         &	Microarray	& ASU			\\
\hline
\end{tabular}
\end{center}
\end{table}

\subsection{Classifiers}

The three classifiers we use as wrappers are described below\footnote{In our experiments, we made use of the corresponding MATLAB implementations for all three classifiers.}. 

\begin{itemize}

\item {\it k-Nearest Neighbors (NN)}: A non-parametric method that classifies a new instance with respect to the majority class labels of its $k$ nearest neighbors. In our experiments, we use the 1-NN method with the usual Euclidean distance as the distance metric.

\item {\it Decision Tree Method (C4.5)}: A technique that uses a tree structure consisting of a root, branches, and intermediate and leaf nodes where the paths from the root to the leaf nodes correspond to classification rules. The particular type we use is the C4.5 implementation~\citep{Quinlan93}.

\item {\it Support Vector Machine (SVM)}: A binary classification technique that performs prediction by finding the maximal margin hyper-planes separating the classes in the feature space~\citep{cortes-1995}. The particular implementation we use is the linear SVM.

\end{itemize}

\subsection{Performance Criterion}

For a specific feature subset of a given dataset, (noisy) loss function measurements are computed as the mean of 10 repeated 5-fold cross-validation (CV) error (i.e., misclassification) rates with respect to that feature subset. In the CV procedure, the dataset is first divided into $k$ random disjoint folds with roughly equal sizes. Next, each one of the folds is used for testing after training the classification algorithm using the remaining $k-1$ folds. The performance of the classifier is then evaluated by averaging these $k$ error rates. In each repetition of the CV procedure, we randomly form new $k$ disjoint folds to compute the CV error rate. The purpose of this approach is to ensure that the particular partitioning in a repetition is an independent sample from the population of all possible ways of $k-$fold partitioning of the original data.

Averaging errors over 10 random repetitions of the CV procedure reduces the variance of noise and makes the FS process more robust to the variability in the CV process. For instance, it could happen that a single fold contains mostly a specific class and the others not, so averaging helps in dealing with such scenarios. This convention for loss function evaluation is applied uniformly and consistently throughout our computational experiments for all FS methods. Our choice of this criterion is due to the fact that CV has proved to be one of the most popular and widely-used performance evaluation methods in pattern recognition~\citep{Wong15}.

\subsection{Method Parameters}
\label{sec:methlgy}

Careful fine-tuning of BSPSA parameters is critical for convergence of the algorithm to a good solution. During the fine-tuning process, we observed that using two different sets of parameters, one for small and one for large datasets, exhibited better convergence properties compared to a single set of parameters. This fine-tuning process is described below.

A simple, theoretically valid, and commonly used distribution for each component of the simultaneous perturbation vector $\Delta_k$ is the Bernoulli distribution with a probability of 0.5 for each $\pm$1 outcome. Our BSPSA implementation as well as CSPSA make use of this distribution for generating $\Delta_k$.

For each BSPSA run, each component of the initial estimate $w_0$ was set to 0.5. The stability constant $A$ was taken as 100 and 300 respectively for small and large datasets per~\citet{Spall98_2}, which recommends the stability constant be at least 10\% of the number of iterations.

\begin{figure}[t]
\centering
\includegraphics[width = 3.5in]{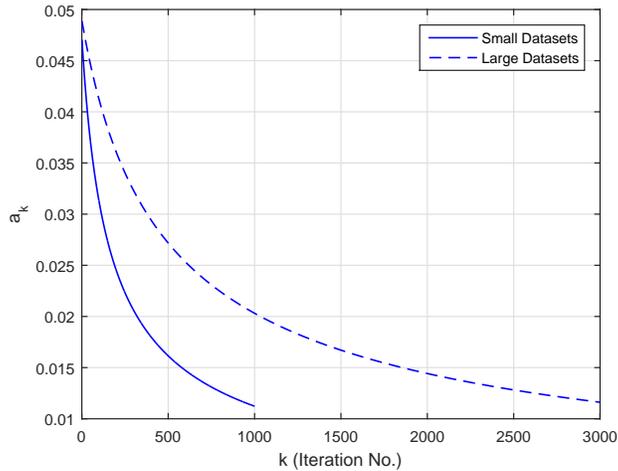}
\caption{Gain sequence $a_k$ for small and large datasets}
\label{fig:gain}
\end{figure}

In our tests, we observed that a perturbation of 0.05 units in the elements of $\hat{w}_k$ was usually sufficient for acceptable SPSA performance, so $c$ was taken as 0.05. That is, $\hat{w}_k^{\pm} = \hat{w}_k {\pm} (0.05) \Delta_k$. \citet{Spall92} notes the optimal value for $\alpha$ in a finite-sample setting is 0.6, which we found to work well in our experiments. Once $A$, $c$ and $\alpha$ were decided upon, we chose $a$ such that the change in the magnitude of $w_k$ components in early iterations is about half the difference between $y(\hat{w}_k^{+})$ and $y(\hat{w}_k^{-})$ (after bounding and rounding of $\hat{w}_k^{\pm}$). For small datasets, $a$ was taken as 0.75 whereas for large datasets, it was set to 1.5. For instance, suppose that $y(\hat{w}_k^{+})=0.12$ and $y(\hat{w}_k^{-})=0.08$ as in Figure~\ref{fig:block} with a difference of 0.04. Note that $a/(A+1)^\alpha = 0.047$, which results in a 0.19 magnitude change in the components of $w_k$. The $a_k$ gain sequences for both dataset types are shown in Figure~\ref{fig:gain}.

For CSPSA, we used the implementation of~\citet{JohannsenWSP04}. Both BSPSA and CSPSA were allowed to run for a maximum of 1000 iterations for small datasets and 3000 iterations for the large ones. BSPSA and CSPSA parameters are summarized in Table~\ref{tab:SPSA_par}.

We now briefly discuss the BGA implementation of~\citet{OluleyeALD14}. This implementation starts with chromosomes that are random $p-$dimensional bit strings. As for the selection scheme, a tournament of size 2 is employed with an elite count of 2 chromosomes. The mutation operator used is the uniform bit flipping with a 0.1 probability. As for crossover, the arithmetic crossover is used wherein two parents are combined via the XOR operator with the crossover probability taken as 0.8. Recall that both BSPSA and CSPSA require three objective function evaluations in each iteration. For a fair comparison, BGA was allowed to run with the same total number of maximum objective function evaluations as in BSPSA and CSPSA. Specifically, for the small datasets, BGA population size was set to 30 with a maximum of 100 generations whereas for the large datasets, BGA population size was taken as 45 with a maximum of 200 generations. This way, all three methods had a budget of 3000 function evaluations for the small datasets and a budget of 9000 function evaluations for the large ones. BGA parameters are tabulated in Table~\ref{tab:BGA_par}.

In case the objective function value did not improve for more than 25\% of the maximum number of iterations (or generations in BGA), the run was declared to have stalled and it was terminated. This corresponded to 250 iterations for small datasets and 750 iterations for the large ones. For BGA, the stall limit corresponded to 25 and 50 generations respectively.

BSPSA, CSPSA, and BGA were coded in MATLAB and the experiments were conducted on a workstation with 16 CPU cores and 3.1 GHz clock speed. For shorter execution times, the 10 cross-validation repetitions were carried out in parallel.

\begin{table}
\centering
\fontsize{8}{11pt}\selectfont
\caption{BSPSA and CSPSA parameters}\label{tab:SPSA_par}
  \begin{tabular}{lllll}
    \toprule
    \multirow{2}{*}{Parameter} &
      \multicolumn{2}{c}{BSPSA} &
      \multicolumn{2}{c}{CSPSA} \\
      & {\shortstack{Small\\Dataset}} & {\shortstack{Large\\Dataset}} & {\shortstack{Small\\Dataset}} & {\shortstack{Large\\Dataset}} \\
      \midrule
Max no. of Iterations	& 1000	& 3000	& 1000	& 3000 	\\
No. of Stall Iterations & 250   & 250   & 750   & 750 	\\
$c$						& 0.05	& 0.05	& 0.01	& 0.01	\\
$a$						& 0.75	& 1.5	& 0.75	& 0.75	\\
$A$						& 100	& 300	& 0		& 0		\\
$\alpha$				& 0.6	& 0.6	& 0.6	& 0.6	\\
$\gamma$				& N/A	& N/A	& 0.1	& 0.1	\\
Components of $w_0$		& 0.5	& 0.5	& 0.5	& 0.5	\\
    \bottomrule
  \end{tabular}
\end{table}

\begin{table}
\fontsize{8}{11pt}\selectfont
\begin{center}
\caption{BGA parameters}\label{tab:BGA_par}
\begin{tabular}{l l l}\hline
Parameter           & Small Datasets    & Large Datasets\\
\hline
Population Size		& 30				& 45			\\
Max. No. of Generations	& 100      		& 200	    \\
No. of Stall Generations & 25        & 50  	\\
Elite Count			& 2      		& 2	        \\
Mutation Probability   	& 0.1       		& 0.1	    \\
Crossover Probability     & 0.8            & 0.8	    \\
\hline
\end{tabular}
\end{center}
\end{table}

\subsection{Results}

As mentioned earlier, a function measurement is taken as the average of 10-repeated 5-fold CV error rates. BSPSA number of iterations and execution times for each dataset are reported in Table~\ref{tab:run_time}.

\begin{table}[t]
\fontsize{8}{11pt}\selectfont
\caption{Number of iterations and execution time of one BSPSA run (in minutes)}\label{tab:run_time}
\begin{center}
\begin{tabular}{lllllll}
\toprule
    \multirow{2}{*}{Dataset} &
      \multicolumn{2}{c}{NN} &
      \multicolumn{2}{c}{C4.5} &
      \multicolumn{2}{c}{Linear-SVM}\\
      & \# Iter. & Time & \# Iter. & Time & \# Iter. & Time \\
      \midrule 
Ionosphere		& 592  & 4.19  & 808  & 3.94 	& 725  & 4.59			\\
Segmentation	& 539  & 1.94  & 559  & 4.87 	& 263  & 9.42			\\
Sonar			& 971  & 3.23  & 465  & 2.39 	& 764  & 4.41			\\
Vehicle			& 429  & 1.52  & 288  & 2.57	& 255  & 3.30			\\
Waveform		& 667  & 2.99  & 425  & 14.35	& 883  & 859.93		\\ 
\hline
Arrhythmia		& 3000 & 8.63  & 2090 & 41.28   & 1304 & 203.57		\\
AR10P			& 3000 & 12.95 & 3000 & 81.72	& 3000 & 706.32		\\
CLL-SUB-111		& 2186 & 14.72 & 2773 & 133.92  & 3000 & 184.92		\\
GLI-85			& 1624 & 16.09 & 2099 & 80.73   & 3000 & 126.5			\\
TOX-171			& 3000 & 21.65 & 1823 & 102.70  & 3000 & 276.4			\\ 
\hline
\end{tabular}
\end{center}
\end{table}	

For small datasets, Tables~\ref{tab:FS-small-NN},~\ref{tab:FS-small-DT}, and~\ref{tab:FS-small-SVM} show the averages and standard errors (i.e., standard deviations divided by $\sqrt{10}$) for the 10-repeated 5-fold CV classification error percentages using the classifiers NN, C4.5 and Linear-SVM respectively. The tables show the errors for the full set of features along with the best feature subsets found by BSPSA, CSPSA, BGA, and the three conventional FS methods SFFS, SFS, and SBS. The tables also show the number of features of the best solution found by the respective methods and the number of loss function evaluations.

\begin{table}[ht]
\fontsize{8}{11pt}\selectfont
\begin{center}
\caption{Average CV error percentages of Neighest Neigbor classifier for small datasets along with standard errors, number of selected features, and number of function evaluations. Statistically better methods are indicated with a star.}\label{tab:FS-small-NN}
\begin{tabular}{p{1.5cm} p{0.6cm} p{0.6cm} p{0.6cm} p{0.6cm} p{0.6cm} p{0.6cm} p{0.6cm}}\hline
				& Full	& BSPSA				& CSPSA & BGA 	& SFFS	& SFS	& SBS	\\\hline
Ionosphere	& & & & & & & \\\hline	
Error (\%)			& 13.56 & \textbf{5.80*}	& 9.91  & 8.12	& 7.07	& 8.01	& 11.99	\\
Std. Error 		& 0.18 & 0.15 & 0.19  & 0.14 & 0.25  & 0.22 & 0.26\\
\# Features		& 34 & 11 & 20  & 16  & 8 & 4 & 31\\
\# F-Evals	& 1 & 1776 & 759  & 1410  & 743 & 131 & 191\\
\hline 
Segmentation	& & & & & & & \\\hline 
Error (\%)			& 3.81 & \textbf{2.75}	& 3.34  & 2.87	& 2.84	& 2.94	& 2.96	\\
Std. Error 		& 0.07 & 0.04 & 0.05  & 0.05 & 0.03  & 0.07 & 0.05\\
\# Features		& 19 & 7 & 13  & 12  & 8 & 4 & 16\\
\# F-Evals	& 1 & 1617 & 3000  & 1110  & 517 & 146 & 126\\
\hline 
Sonar	& & & & & & & \\\hline
Error (\%)			& 13.65 & \textbf{4.81*}	& 11.44  & 8.66	& 14.18	& 12.69	& 9.18	\\
Std. Error 		& 0.44 & 0.33 & 0.51  & 0.32 & 0.62  & 0.48 & 0.64\\
\# Features		& 60 & 33 & 26  & 35  & 6 & 8 & 54\\
\# F-Evals	& 1 & 2913 & 1233  & 2130  & 3094 & 556 & 179\\
\hline 
Vehicle	& & & & & & & \\\hline
Error (\%)			& 30.38 & \textbf{26.74}	& 29.02  & 27.78	& 27.71	& 26.88	& 29.11	\\
Std. Error 		& 0.40 & 0.28 & 0.27  & 0.24 & 0.21  & 0.26 & 0.26\\
\# Features		& 18 & 10 & 12  & 8  & 10 & 9 & 16\\
\# F-Evals	& 1 & 1287 & 1452  & 1590  & 307 & 127 & 68\\
\hline 	
Waveform	& & & & & & & \\\hline	
Error (\%)			& 23.68 & 20.30 & 23.46 & \textbf{20.12} & 21.52  & 21.16	& 20.92	\\
Std. Error 		& 0.09 & 0.10 & 0.11  & 0.08 & 0.09  & 0.11 & 0.06\\
\# Features		& 21 & 17 & 21  & 17  & 18 & 15 & 17\\
\# F-Evals	& 1 & 2001 & 2256  & 2010  & 548 & 204 & 97\\
\hline
\end{tabular}
\end{center}
\end{table}

\begin{table}[ht]
\fontsize{8}{11pt}\selectfont
\begin{center}
\caption{Average CV error percentages of Decision Tree classifier for small datasets along with standard errors, number of selected features, and number of function evaluations. Statistically better methods are indicated with a star.}\label{tab:FS-small-DT}
\begin{tabular}{p{1.5cm} p{0.6cm} p{0.6cm} p{0.6cm} p{0.6cm} p{0.6cm} p{0.6cm} p{0.6cm}}\hline
				& Full	& BSPSA				& CSPSA & BGA 	& SFFS	& SFS	& SBS	\\\hline
Ionosphere	& & & & & & & \\\hline	
Error (\%)			& 11.68 & \textbf{6.38*}	& 9.17  & 8.60	& 8.46	& 8.66	& 8.92	\\
Std. Error 		& 0.20 & 0.28 & 0.29  & 0.34 & 0.28  & 0.32 & 0.26\\
\# Features		& 34 & 11 & 13  & 18  & 12 & 6 & 28\\
\# F-Evals	& 1 & 2424 & 1161  & 1440  & 999 & 161 & 132\\
\hline 
Segmentation	& & & & & & & \\\hline 
Error (\%)			& 4.70 & 3.40& 4.73& \textbf{3.37}	& 3.42  & 3.87	& 4.18	\\
Std. Error 		& 0.09 & 0.08 & 0.07  & 0.10 & 0.08  & 0.06 & 0.05\\
\# Features		& 19 & 9 & 19  & 8  & 5 & 6 & 16\\
\# F-Evals	& 1 & 1677 & 762  & 1230  & 344 & 136 & 87\\
\hline 
Sonar	& & & & & & & \\\hline
Error (\%)			& 28.89 & \textbf{19.13}	& 23.85  & 20.96	& 20.58	& 19.57	& 27.74	\\
Std. Error 		& 0.88 & 0.59 & 0.89  & 0.50 & 0.68  & 1.02 & 0.68\\
\# Features		& 60 & 18 & 30  & 19  & 5 & 5 & 59\\
\# F-Evals	& 1 & 1395 & 1530  & 3000  & 739 & 346 & 179\\
\hline 
Vehicle	& & & & & & & \\\hline	
Error (\%)			& 29.21 & \textbf{26.10*} & 29.05 & 28.39 & 29.65  & 29.69	& 28.87	\\
Std. Error 		& 0.43 & 0.35 & 0.39  & 0.30 & 0.30  & 0.49 & 0.40\\
\# Features		& 18 & 15 & 11  & 8  & 17 & 6 & 15\\
\# F-Evals	& 1 & 864 & 1026  & 1740  & 254 & 127 & 95\\
\hline 	
Waveform	& & & & & & & \\\hline	
Error (\%)			& 24.93 & \textbf{23.92} & 24.94 & 24.37 & 24.39  & 24.71	& 24.51	\\
Std. Error 		& 0.22 & 0.14 & 0.16  & 0.15 & 0.18  & 0.20 & 0.11\\
\# Features		& 21 & 16 & 13  & 14  & 12 & 11 & 20\\
\# F-Evals	& 1 & 1275 & 1410  & 2490  & 816 & 204 & 80\\
\hline
\end{tabular}
\end{center}
\end{table}

\begin{table}[tt]
\fontsize{8}{11pt}\selectfont
\begin{center}
\caption{Average CV error percentages of Linear SVM classifier for small datasets along with standard errors, number of selected features, and number of function evaluations. Statistically better methods are indicated with a star.}\label{tab:FS-small-SVM}
\begin{tabular}{p{1.5cm} p{0.6cm} p{0.6cm} p{0.6cm} p{0.6cm} p{0.6cm} p{0.6cm} p{0.6cm}}\hline
				& Full	& BSPSA				& CSPSA & BGA 	& SFFS	& SFS	& SBS	\\\hline
Ionosphere	& & & & & & & \\\hline	
Error (\%)			& 12.36 & \textbf{8.55*}	& 10.77  & 9.18	& 11.62	& 11.62	& 11.74	\\
Std. Error 		& 0.21 & 0.13 & 0.24  & 0.13 & 0.06  & 0.13 & 0.24\\
\# Features		& 34 & 15 & 16  & 19  & 4 & 6 & 32\\
\# F-Evals	& 1 & 2175 & 1047  & 2460  & 2594 & 161 & 387\\
\hline 
Segmentation	& & & & & & & \\\hline 
Error (\%)			& 5.17 & 4.16 & 4.97& \textbf{4.01}	& 4.81  & 4.71	& 4.96	\\
Std. Error 		& 0.05 & 0.03 & 0.06  & 0.05 & 0.05  & 0.03 & 0.03\\
\# Features		& 19 & 7 & 14  & 11  & 10 & 8 & 18\\
\# F-Evals	& 1 & 789 & 825  & 1050  & 446 & 155 & 39\\
\hline 
Sonar	& & & & & & & \\\hline
Error (\%)			& 25.29 & \textbf{13.61*}	& 18.04  & 16.81	& 19.76	& 21.92	& 24.86	\\
Std. Error 		& 0.48 & 0.41 & 0.27  & 0.55 & 0.32  & 0.43 & 0.37\\
\# Features		& 60 & 25 & 23  & 25  & 9 & 4 & 58\\
\# F-Evals	& 1 & 2292 & 2262  & 2580  & 730 & 235 & 121\\
\hline 
Vehicle	& & & & & & & \\\hline
Error (\%)			& 20.21 & 19.32& 20.02& 19.17 & \textbf{18.98}	& 23.56  & 19.09	\\
Std. Error 		& 0.26 & 0.13 & 0.12  & 0.23 & 0.20  & 0.15 & 0.16\\
\# Features		& 18 & 18 & 18  & 15  & 15 & 10 & 14\\
\# F-Evals	& 1 & 765 & 858  & 1290  & 826 & 171 & 68\\
\hline 	
Waveform	& & & & & & & \\\hline	
Error (\%)		& 13.09 & 12.96& 13.85& \textbf{12.18} & 13.12 & 12.98 & 13.09	\\
Std. Error 		& 0.04 & 0.05 & 0.05  & 0.06 & 0.04  & 0.04 & 0.05\\
\# Features		& 21 & 18 & 19  & 18  & 20 & 18 & 19\\
\# F-Evals	& 1 & 2649 & 1377  & 1230  & 559 & 226 & 43\\
\hline
\end{tabular}
\end{center}
\end{table}

At this point, a few remarks are in order on statistical tests regarding the CV error, which remains an active area of research. In the case of repeated CV procedures, the error rate is a population mean and not a proportion, and it can be approximated by a normal distribution~\citep{zha15}.~\citet{Wong15} advises against use of repetition in CV due to (weak) statistical dependency between the repetitions. However, it is quite difficult to arrive at statistically significant results with a single CV error measurement as the variance at the fold level can be rather drastic (as high as 25\% in the datasets we consider). In addition, a universal unbiased estimator of $k$-fold CV does not exist~\citep{ben04}, which further complicates theoretical validity of any potential conclusions. 

Nonetheless, the primary statistical analysis procedure we employ in this study is one-way ANOVA test on the mean CV error for each dataset/ classifier combination. All statistical tests were conducted at a 5\% significant level. We first used the Bartlett's test to check if the FS methods have equal group variance for a given combination. In the case of statistically significant unequal variances (which was the case in only two combinations), we proceeded after a Welch correction for variance nonhomogeneity. If the ANOVA test indicated a statistical difference in the mean CV errors, we conducted pair-wise comparisons using Tukey's HSD test. If this test indicated that any one method is statistically better than the others, we indicated this with a star for the corresponding method. 

As can be seen in Tables~\ref{tab:FS-small-NN},~\ref{tab:FS-small-DT}, and~\ref{tab:FS-small-SVM}, for small datasets, out of the 15 dataset/ classifier combinations, six combinations have statistically better FS results, all of which belong to BSPSA. In all but five combinations, BSPSA gives the lowest mean error, and for these five combinations where BSPSA is outperformed, the difference is not statistically significant. 

Table~\ref{tab:FS-large} shows FS results for the large datasets with respect to the full set of features and those obtained by BSPSA, CSPSA, and BGA. In the table, the standard errors (i.e., standard deviations divided by $\sqrt{10}$) are for the 10-repeated 5-fold CV classification error percentages. Except for the Arrhythmia dataset where the difference between the FS methods was not statistically significant, BSPSA identified feature sets that were statistically better in all dataset/ classifier combinations. CSPSA stalled before reaching the maximum number of iterations in all of the 15 dataset/ classifier combinations and yielded features that were inferior compared to BSPSA in general. As for BGA, it was outperformed by BSPSA by a large margin in most cases. It appeared that either the allowed number of generations was not adequate for BGA for convergence or, in about half the cases, BGA stalled before reaching the maximum number of generations.

We observed that BSPSA cuts the mean CV error by about half on the average, and as much as 15 fold as in the case of the AR10P and Linear-SVM combination. With the NN classifier, for the AR10P image dataset with 2400 features, FS with BSPSA resulted in almost a 52\% decrease in average CV error in less than 13 minutes. For the GLI-85 microarray dataset with more than 22000 features, BSPSA was able to reduce the average CV error from 13.29\% to 7.51\% in about 15 minutes. As for TOX-171, BSPSA reduced the error rate from 14.21\% down to 3.98\% in about 20 minutes, providing a $(14.21-3.98)/14.21*100=72\%$ reduction in the average CV error rate.

\begin{table}
\centering
\fontsize{8}{11pt}\selectfont
\caption{Average CV error percentages for large datasets along with standard errors, number of selected features, and number of function evaluations. Statistically better methods are indicated with a star.}\label{tab:FS-large}
  \begin{tabular}{lllllllllllll}
    \toprule
    \multirow{2}{*}{} &
      \multicolumn{4}{c}{NN} &
      \multicolumn{4}{c}{C4.5} &
      \multicolumn{4}{c}{Linear-SVM} \\
    & {Full} & {BSPSA} & {CSPSA} & {BGA} & {Full} & {BSPSA} & {CSPSA} & {BGA} & {Full} & {BSPSA} & {CSPSA} & {BGA} \\
      \midrule
Arrhythmia & & & & & & & & & & & &\\\hline
Error (\%) & 52.94 & 29.60 & 38.1 & \textbf{28.89} & 35.46 & \textbf{28.70} & 33.97 & 28.94 & 41.42 & 29.51 & 31.50 & \textbf{28.76}\\ 
Std. Error  & 0.85 & 0.36 & 0.32 & 0.24 & 0.43 & 0.25 & 0.34 & 0.36 & 0.13 & 0.17 & 0.22 & 0.15 \\
\# Features& 279 & 133 & 134& 176 & 279 & 132 & 131 & 158 & 279 & 149 & 124 & 156\\
\# F-Evals & 1 & 9000 & 2343 & 9000 & 1 & 6270 & 3234 & 9000 & 1 & 3912 & 2667 & 7920 \\			
\hline
AR10P		& & & & & & & & & & & &\\\hline
Error (\%) & 47.08 & \textbf{22.80*} & 43.46 & 36.62 & 31.00 & \textbf{15.15*} & 22.73 & 23.26 & 3.31 & \textbf{0.21*} & 1.69 & 1.46\\
Std. Error  & 0.58 & 0.49 & 0.49 & 0.65 & 1.15 & 0.87 & 0.92 & 0.96 & 0.41 & 0.15 & 0.27 & 0.35\\
\# Features& 2400 & 1096 & 1192 & 1237 & 2400 & 1118 & 1218 & 1182 & 2400 & 1123 & 1130 & 1186\\
\# F-Evals & 1 & 9000 & 4716 & 9000 & 1 & 9000 & 4485 & 5895 & 1 & 9000 & 4179 & 2925\\
\hline
CLL-SUB-111	& & & & & & & & & & & &\\\hline
Error (\%) & 40.36 & \textbf{23.87*} & 33.56 & 28.29 & 39.10 & \textbf{26.03*} & 30.41 & 31.29 & 22.88 & \textbf{9.31*} & 17.48 & 12.44\\
Std. Error  & 1.07 & 0.57 & 0.59 & 0.72 & 1.26 & 1.17 & 1.30 & 1.55 & 0.71 & 0.47 & 0.48 & 0.77 \\ 
\# Features& 11340 & 5647 & 5718 & 6208 & 11340 & 5672 & 5646 & 5679 & 11340 & 5673 & 5742 & 5701\\
\# F-Evals & 1 & 6558 & 5058 & 9000 & 1 & 8319 & 2268 & 6030 & 1& 9000 & 3711 & 9000\\
\hline
GLI-85 & & & & & & & & & & & &\\\hline
Error (\%) & 13.29 & \textbf{7.51*} & 10.01 & 10.47 & 21.41 & \textbf{10.35*} & 12.94 & 17.65 & 9.53 & \textbf{6.54*} & 8.47 & 7.88\\
Std. Error  & 0.78 & 0.33 & 0.48 & 0.62 & 1.20 & 0.60 & 0.43 & 0.81 & 0.48 & 0.18 & 0.41 & 0.25\\
\# Features& 22283 & 11271 & 11197 & 11591 & 22283 & 11152 & 11188 & 11506 &22283 & 11023 & 11006 & 11354\\
\# F-Evals & 1 & 4872 & 2490 & 2880 & 1 & 6297 & 5088 & 3780 & 1 &9000 & 2259 & 5220\\
\hline
TOX-171 & & & & & & & & & & & &\\\hline
Error (\%) & 14.21 & \textbf{3.98*} & 9.94 & 7.37 & 42.51 & \textbf{28.14*} & 32.70 & 34.80 & 4.21 & \textbf{0.57*} & 3.51 & 3.01\\
Std. Error  & 0.85 & 0.35 & 0.45 & 0.43 & 1.07 & 1.03 & 0.86 & 1.74 & 0.41 & 0.23 & 0.20 & 0.31\\
\# Features& 5748 & 2875 & 2887 & 3184 & 5748 & 2802 & 2871 & 3116 & 5748 & 2803 & 2903 & 2954\\
\# F-Evals & 1 & 9000 & 3303 & 4680 & 1 & 5469 & 4053 & 9000 & 1 & 9000 & 3792 & 9000\\
    \bottomrule
  \end{tabular}
\end{table}

Table~\ref{tab:FS-summary} shows the average percent reduction in the CV error rate achieved by BSPSA and BGA for small and large datasets for each classifier. For large datasets, BSPSA achieves a drastic reduction of 49\% in the CV error rate whereas BGA achieves only a 29\% reduction. Across all the experiments including small and large datasets, BSPSA provides a 38\% reduction on the average for the CV error rate, outperforming BGA that yields a 25\% reduction. 

\begin{table}
\centering
\fontsize{8}{11pt}\selectfont
\caption{Average percentage points reduction in CV error rates}\label{tab:FS-summary}
  \begin{tabular}{lllllll}
    \toprule
    \multirow{2}{*}{} &
      \multicolumn{2}{c}{Small Datasets} &
      \multicolumn{2}{c}{Large Datasets} &
      \multicolumn{2}{c}{All Datasets} \\
      & {BSPSA} & {BGA} & {BSPSA} & {BGA} & {BSPSA} & {BGA} \\
      \midrule
NN				& 35.21 & 24.99	& 50.40	& 33.38 & 42.81	& 29.19		\\
C4.5			& 24.30 & 17.43 & 37.82 & 19.81 & 31.06 & 18.62		\\
Linear-SVM		& 20.39 & 18.76 & 59.91 & 35.58 & 40.15 & 27.17		\\ \hline
Average			& 26.63 & 20.39 & 49.38 & 29.59 & 38.01 & 25.00		\\
    \bottomrule
  \end{tabular}
\end{table}

\begin{figure}
\captionsetup[subfigure]{labelformat=empty}
\begin{center}
\subfloat[]{\includegraphics[width = 4in]{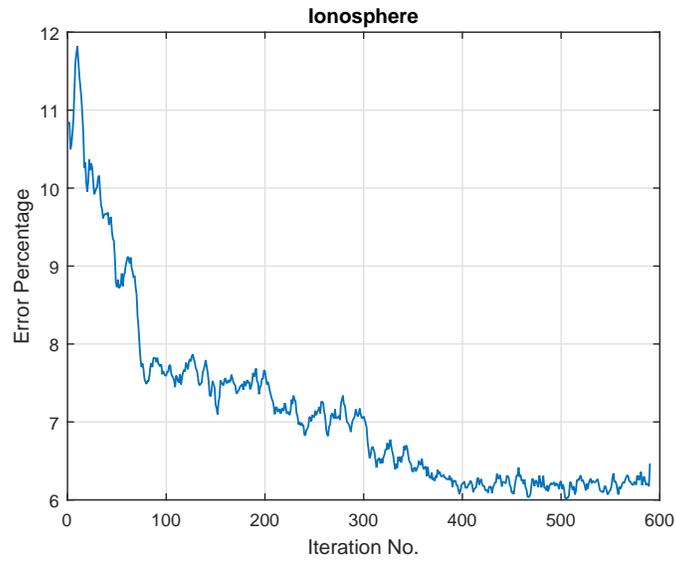}}\\ 
\vspace{-.7cm}
\caption{Error percentage vs. iteration for the Ionosphere dataset with the NN classifier}
\label{error_small}
\end{center}
\end{figure}

\begin{figure}
\captionsetup[subfigure]{labelformat=empty}
\begin{center}
\subfloat[]{\includegraphics[width = 4in]{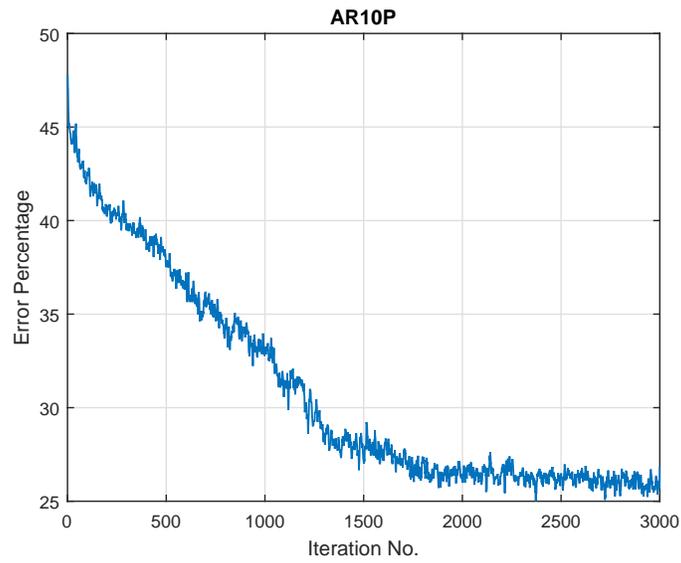}}\\
\vspace{-.7cm}
\caption{Error percentage vs. iteration for the AR10P dataset with the NN classifier}
\label{error_large}
\end{center}
\end{figure}

Figures~\ref{error_small} and \ref{error_large} show the average CV error percentage for the feature subset found by BSPSA at each iteration for the Ionosphere and AR10P datasets respectively with the NN classifier after a moving average smoothing. For small datasets, we observed that BSPSA can find very good solutions even within the first several hundred iterations. For the large datasets, our observation was that BSPSA slowly and steadily converges to a good solution in general. 

In our last set of experiments, we compared BSPSA against the full set of features and BGA with the purpose of maximizing the area under the ROC curve (AUC) metric. This metric was computed as the average of 10 repeated measurements of 5-fold cross-validated AUC. In this comparison, we used the Ionosphere and Sonar datasets that have 2 classes for the response variable and we experimented with all the three wrapper classifiers. Feature sets found by BSPSA were better than those of BGA in all six dataset/ classifier combinations, with the differences being statistically significant in three combinations. The average AUC comparison results for the NN and C4.5 classifiers are shown in Table~\ref{tab:AUC}.

\begin{table}
\centering
\fontsize{8}{11pt}\selectfont
\caption{Average cross-validated AUC for Ionosphere and Sonar datasets with NN and C4.5 classifiers. Statistically better methods are indicated with a star.}\label{tab:AUC}
  \begin{tabular}{l l l l l l l }
    \toprule
    \multirow{2}{*}{} &
      \multicolumn{3}{c}{NN} &
      \multicolumn{3}{c}{C4.5} \\
      & {Full} & {BSPSA} & {BGA} & {Full} & {BSPSA} & {BGA}  \\
      \midrule
Ionosphere		&  	&  	&  	&  	&  	&  \\ \hline
AUC				& 0.813 & \textbf{0.934*} & 0.911  & 0.802	& \textbf{0.952}& 0.941 \\
Std. Error 		& 0.007 & 0.007 & 0.006  & 0.009 & 0.010  & 0.012 \\
\# Features		& 34 & 11 & 7  & 34  & 18 & 9 \\
\# F-Evals		& 1 & 2322 & 3000  & 1 & 1281 & 3000\\
\hline 
Sonar			&  	&  	&  	&  	&  	&  \\ \hline
AUC				& 0.866 & \textbf{0.964*} & 0.899  & 0.713	& \textbf{0.822*}& 0.787	\\
Std. Error 		& 0.008 & 0.009 & 0.007  & 0.013 & 0.012  & 0.010 \\
\# Features		& 60 & 27 & 54  & 60  & 19 & 16 \\
\# F-Evals		& 1 & 2520 & 3000  & 1 & 1440 & 3000\\
    \bottomrule
  \end{tabular}
\end{table}

\section{Summary and Conclusions}
\label{sec:conclsn}

Binary SPSA is an attractive method for feature selection due to the following reasons: (1) it does not require an explicit objective function nor its derivatives, making a broad range of classifier performance metrics a feasible choice for the objective function, (2) it formally accounts for noise in evaluation of the performance metric, and (3) it is consistent with the binary nature of the FS problem. In this work, we present practical implementation details and propose specific parameter values for an efficient BSPSA approach to FS. In addition, we provide extensive computational experiments comparing BSPSA against CSPSA, binary GA, and popular conventional FS methods SFFS, SFS, and SBS using NN, C4.5, and Linear-SVM classifiers as wrappers. In our experiments, we use standard benchmark datasets using the average of 10-repeated 5-fold cross-validation error rate as our performance metric. We also perform limited experiments with the average of 10-repeated 5-fold cross-validated AUC.

For small datasets (with less than 100 features), BSPSA solutions compare quite favorably to other FS methods across all dataset/ classifier combinations, either outperforming all of them or yielding comparable results. For large datasets, BSPSA cuts the CV errors almost by half on the average while comfortably outperforming CSPSA and BGA by a large margin in most cases. We are not aware of any wrapper FS method in the literature that can be used effectively on datasets with tens of thousands of features within reasonable execution times and with good convergence properties. 

The fact that BSPSA approximates the gradient only through noisy function measurements gives rise to a vast number of possibilities in feature selection. In particular, it is straightforward to use BSPSA for optimization of various other metrics such as cost-sensitive classification accuracy, root mean square error (RMSE), or error rates of hold-out or leave-one-out methods. It is also straightforward to use BSPSA for FS in regression problems where the response variable is numeric. In this regard, we believe that BSPSA holds a strong potential in tackling difficult feature selection problems.

One characteristic of BSPSA that is especially relevant for large datasets is that, as can be seen in Table~\ref{tab:FS-large}, number of features selected is usually close to half of the available features. The reason for this kind of behavior is that we start with an initial solution vector with half of the available features and, 3000 iterations (which correspond to an extremely small fraction of the solution space) is simply not sufficient for BSPSA to move to solutions with significant deviations in the number of selected features. To address this issue, one might perform sequential BSPSA runs, feeding output of one BSPSA run as an input to the next run, each time reducing the number of selected features by about half, until no more performance gains can be achieved. Nonetheless, this BSPSA characteristic can be used to control the number of features desired in the final output. It might also be helpful to use an appropriate filter FS method prior to running BSPSA to allow the algorithm start with features that already have some level of association with the response variable in order for BSPSA to converge to a good solution with a relatively small number of iterations.

As for future work, one obvious direction is better fine-tuning of the BSPSA parameters. Even though the parameter values we recommend give good results for the datasets we consider, it is plausible that better results can be obtained by tailoring BSPSA parameters for the particular dataset and classifier at hand. Future work might also consider extensions to basic BSPSA for more effective FS: (1) BSPSA might be converted to a global minimizer (in probability) by the technique of injecting noise in the solution vector update step~\citep{mar08}, and (2) second order derivative information can be included for the purpose of constructing a stochastic analogue to the (deterministic) Newton's method for faster convergence~\citep{spall00}.

\section*{Acknowledgments}
The authors are grateful to the associate editor and the three anonymous reviewers for their invaluable comments and suggestions that greatly improved the overall value and quality of this manuscript. This work was supported by The Scientific and Technological Research Council of Turkey (TUBITAK), Grant No. 113M489. The authors thank Prof. James C. Spall with Johns Hopkins University for several helpful discussions. MATLAB source code for BSPSA can be found at GitHub at the following link:\\
\url{http://github.com/vaksakalli/fs_bspsa}


\bibliographystyle{model2-names}
\bibliography{refs}


\end{document}